\documentclass{article}

\usepackage{PRIMEarxiv}

\usepackage[utf8]{inputenc} 
\usepackage[T1]{fontenc}    
\usepackage{hyperref}       
\usepackage{url}            
\usepackage{booktabs}       
\usepackage{amsfonts}       
\usepackage{nicefrac}       
\usepackage[verbose=true]{microtype}     
\usepackage{lipsum}
\usepackage{fancyhdr}       
\usepackage{graphicx}       

\usepackage[numbers]{natbib}
\usepackage{kotex}
\usepackage{tipa}
\usepackage{authblk}

\graphicspath{{media/}}     

\title{Automatic Speech Recognition (ASR) for the Diagnosis of pronunciation of Speech Sound Disorders in Korean children
\thanks{Preprint. Under review.}
}

\author[1,2]{\textbf{Taekyung Ahn}}
\author[2]{\textbf{Yeonjung Hong}}
\author[2]{\textbf{Younggon Im}}
\author[3]{\textbf{Do Hyung Kim}}
\author[3]{\textbf{Dayoung Kang}}
\author[3]{\textbf{Joo Won Jeong}}
\author[3]{\textbf{Jae Won Kim}}
\author[4]{\textbf{Min Jung Kim}}
\author[5]{\textbf{Ah-ra Cho}}
\author[3,$\dagger$]{\textbf{Dae-Hyun Jang}}
\author[1,$\dagger$]{\textbf{Hosung Nam}}

\affil[1]{Department of English Language and Literature, Korea University, Seoul, Republic of Korea}
\affil[2]{MediaZen, Seongnam-si, Republic of Korea}
\affil[3]{Incheon St.Mary's Hospital, College of Medicine, The Catholic University of Korea, Incheon, Republic of  Korea}
\affil[4]{Department of Special Education, Dankook University, Youngin-si, Republic of Korea}
\affil[5]{Department of Rehabilitation Medicine, Eunpyeong St.Mary's Hospital, College of Medicine, The Catholic University of Korea, Seoul, Republic of  Korea}

\affil[$\dagger$]{Corresponding author: dhjangmd@naver.com, hnam@korea.ac.kr}

\begin{document}
\maketitle

\begin{abstract}
This study presents a model of automatic speech recognition (ASR) that is designed to diagnose pronunciation issues in children with speech sound disorders (SSDs) to replace manual transcriptions in clinical procedures. Because ASR models trained for general goals mainly predict input speech into real words, it is not possible to use a well-known high-performance ASR model for the purposes of the evaluation of pronunciation for children with SSDs. We fine-tuned the wav2vec2.0 XLS-R model to recognise speech as pronounced rather than as existing words. The model was fine-tuned with a speech dataset of 137 children with inadequate speech production pronouncing 73 Korean words that are selected for actual clinical diagnosis. The model’s predictions of the pronunciations of the words matched the human annotations with about 90\% accuracy. While the model still requires improvement in terms of the recognition of unclear pronunciation, this study demonstrates that ASR models can streamline complex pronunciation error diagnostic procedures in clinical fields.
\end{abstract}


\section{Introduction}
Generally, children with speech sound disorders (SSDs) are clinically diagnosed by speech-language pathologists who transcribe the child’s speech and manually analyse pronunciation errors. Although there have been several attempts to automatically analyse pronunciation errors in child SSD speech \cite{dudy2015pronunciation,dudy2018automatic}, traditional automatic speech recognition (ASR) training methods were unable to achieve the desired recognition performance to replace human annotations. Training traditional ASR models requires a substantial amount of accurately annotated speech data. In addition, traditional ASR models require a pronunciation dictionary called a lexicon for combining in an acoustic model with learned speech features via phonetic symbols. A language model (LM) that has computed the probability of word chains based on correct grammar and vocabulary is also required \cite{povey2011kaldi}. For child SSD speech data, it is difficult and time-consuming not only to gather a transcribed speech but also to build hand-designed pronunciation dictionaries with several pronunciations having the same spelling.

However, over a few short years, the development of the end-to-end-based (e2e-based) model training method generated new possibilities in the spectrum of ASR \cite{chorowski2015attention,bahdanau2016end}. Initially, e2e modelling required a larger dataset than traditional ASR for learning both acoustic and linguistic features, instead of the use of complex processes and pronunciation dictionaries \cite{chiu2018state}. However, large pre-trained models that already have learned a range of speech expressions began to be released in open-source platforms and several studies demonstrated sufficient recognition performance through fine-tuning them without being required to an ASR model from scratch \cite{abate2021end,zuluaga2023pre,xu2022simple,yi2021applying}. These studies provide innovative means of solving the data scarcity problem for downstream tasks in speech recognition.

Taking advantage of this approach, studies have been conducted to develop e2e-based ASR models for the detection of mispronunciation as a downstream task \cite{feng2020sed}. The main goal of mispronunciation detection is to identify pronunciation errors within speech with the use of an ASR model. Because the task of mispronunciation detection requires learning features of speech that have huge acoustic and linguistic variation, the task of the detection of mispronunciation posed several limitations for training an ASR model in typical way. However, following the emergence of an e2e-based ASR network, studies using e2e models have demonstrated that such a model only requires a small amount of data for the task of the detection of mispronunciation with fine-tuning \cite{liu2023automatic,peng2021study,peng2023end,yang2022improving}. The detection of mispronunciation is primarily studied to recognise pronunciation errors in English by non-native English learners to improve second language learning, but thanks to the advancement of ASR technologies, it is also being used in research to recognise the pronunciation error of SSDs for treatment \cite{lee2022end,escobargrisales2023deep} according to various e2e-based ASR models.

Against this background, we introduce an e2e-based ASR model for diagnosing pronunciation errors in children with SSDs in Korea for clinical assessment. Our model was trained to identify pronunciation variation in 73 specific words used for diagnosis in actual clinical practice. Narrowing down the scope of words for recognition, we can improve the accuracy of the detection of pronunciation errors detection without the collection of massive training data.

\section{Related works}
\label{sec:headings}


\subsection{Wav2vec2.0 and XLS-R}
Wav2vec2.0 is a representative e2e-based ASR framework \cite{baevski2020wav2vec}. This training network incorporates self-supervised learning to learn identify the contextual representations of acoustic features. For this method, the main advantage is that the network can learn the context through encoding the raw waveform directly in training, such that no label text corresponding to speech data is required to create a pre-trained model. This network can achieve scores of high accuracy in speech recognition with the use of a limited amount of supervised data for fine-tuning. XLS-R, a cross-lingual pre-trained model that is based on wav2vec2.0 effectively capitalises on key elements in this framework. The model was pre-trained using wav2vec2.0 with large unlabelled 128 languages \cite{conneau2021unsupervised,babu2022xls}, allowing for the construction of a speech recognition model for ultra-low resources and for domain-specific conditions \cite{baevski2020wav2vec} through fine-tuning.

Recent studies have adopted wav2vec2.0 for phoneme recognition and mispronunciation detection, comparing the fine-tuning performance of English-only and multilingual pre-trained models \cite{xu2022simple,zuluaga2023pre,peng2021study}. Such works have shown that the XLS-R model has improved recognition performance for mispronunciation detection tasks. Another study demonstrated that it is possible to create a pronunciation-assessment model for children with SSDs through fine-tuning of the wav2vec2.0 pre-trained model \cite{getman2022wav2vec2}. Following these studies, in this work, we fine-tuned the wav2vec2.0 XLS-R model to create a speech recognition model for this downstream task.

As with traditional ASR modelling, LMs can be used for the decoding of the output of e2e-based ASR models to improve their recognition performance. LMs receiving additional weighting when they successfully predict a text. For the wav2vec2.0 framework, a study improved recognition performance by using LM weighting in downstream tasks \cite{zuluaga2023pre}. Another study showed better performance with LMs for low-resource speech recognition tasks \cite{yi2021applying}. We considered our study to have similar tasks to the studies noted above as its final goal was to analyse the type of pronunciation errors for the diagnosis of SSDs based on the recognition results of given words. In this context, we investigated whether the model performance could be further improved using LM weights to predict cases of pronunciation error in the speech of children with SSDs (https://huggingface.co/blog/wav2vec2-with-ngram).



\subsection{Whisper}
Whisper, a state-of-the-art ASR model, achieved better performance than wav2vec2.0 on an English dataset through training on 680,000 hours of labelled data \cite{radford2022robust}. The huge amount of labelled data used as training data allowed the model to grammatically decode and lexically correct sentences without combining LMs. The training data included 117,000 hours of data in 96 languages, allowing the training model to be used in multilingual recognition tasks. The authors of that study claimed that it was not necessary to fine-tune the model, due to the sufficiently large set of labelled training data. A recent study was conducted to fine-tune Whisper and compared its performance in the recognition of child speech to fine-tuned wav2vec2.0 \cite{jain2023adaptation}. That study found that wav2vec2.0 recognised child speech better than Whisper in task-specific applications. However, this model has not been used in the recognition of the speech of children with SSDs.

For this study, we used Whisper as a baseline model, without the use of fine-tuning. This model is available via \href{https://github.com/openai/whisper}{GitHub}. Whisper is a breakthrough ASR model in recent years. The test results reported in the paper show that the Whisper-large and Whisper-large-v2 models had WERs of 15.2\% and 14.3\% in a Korean speech recognition task. However, we predict that it is still difficult for this well-performing multilingual model to recognise the speech of children with SSDs. In this context, we investigated the need for model building through the comparison of the recognition performance of pre-trained Whisper and wav2vec2.0-based fine-tuned models on the same data.

\section{Materials and methods}
\subsection{Data preparation}

\begin{table}[h]
\renewcommand{\arraystretch}{1.2}
\centering
\begin{tabular}{ll}
\hline
\textbf{Type}                                      & \textbf{Words} \\
\hline
Assessment of Phonology and                        & 거북이[k{\textturnv}bugi], 고래[kor\textipa{E}], 그네[k{\textturnm}ne], 꽃                                                    [k'ot{\textcorner}], 나무[namu],\\
Articulation for Children \cite{kim2007assessment} & 눈사람[nuns'aram], 단추[tants\super{h}u], 딸기[t'algi], 머리                                                                  [m{\textturnv}ri],\\
                                                   & 모자[modza], 바퀴[p\textsubbar{a}\textipa{k\super{h}\textturnh}i], 뱀[p\textipa{E}m], 병원[py{\textturnv}\textipa{N}w{\textturnv}n], 빗[pit{\textcorner}],\\
                                                   & 빨대[p'ald\textipa{E}], 사탕[s\super{h}at\super{h}a\textipa{N}], 색종이[s\super{h}\textipa{E}k{\textcorner}ts'o\textipa{N}i], 시소[{\textctc}is\super{h}o],\\
                                                   & 싸워[s'aw{\textturnv}], 아파[ap\super{h}a], 안경[angy{\textturnv}\textipa{N}], 양말[ya\textipa{N}mal], 없어[{\textturnv}p{\textcorner}s'{\textturnv}],\\
                                                   & 옥수수[ok{\textcorner}s'us\super{h}u], 올라가[ollaga], 우산[us\super{h}an], 이빨[ip'al],\\
                                                   & 장갑[dza\textipa{N}gap{\textcorner}], 찢어[ts'idz{\textturnv}], 책[ts\super{h}\textipa{E}k{\textcorner}], 침대[ts\super{h}imd\textipa{E}], \\
                                                   & 컵[k\super{h}{\textturnv}p{\textcorner}], 토끼[t\super{h}ok'i], 포도[podo], 햄버거[h\textipa{E}mb{\textturnv}g{\textturnv}], \\
                                                   & 호랑이[hora\textipa{N}i], 화장실[hwadza\textipa{N}{\textctc}il] \\
\hline
Urimal Test of Articulation                        & 가방[gaba\textipa{N}], 괴물[kwemul], 귀[kwi], 그네[k{\textturnm}ne], 그림                                                     [k{\textturnm}rim], \\
and Phonology \cite{kim2004urimal}                 & 꼬리[k'ori], 나무[namu], 눈썹[nuns'{\textturnv}p{\textcorner}], 단추                                                          [tants\super{h}u], \\
                                                   & 동물원[do\textipa{N}murw{\textturnv}n], 땅콩[t'a\textipa{N}k\super{h}o\textipa{N}], 로봇[robot{\textcorner}], 메뚜기[met'ugi],\\
                                                   & 못[mot{\textcorner}], 바지[padzi], 뽀뽀[p'op'o], 사탕[s\super{h}at\super{h}a\textipa{N}],\\
                                                   & 세마리[s\super{h}e mari], 싸움[s'aum], 엄마[{\textturnv}mma], 연필[y{\textturnv}np\super{h}il],\\
                                                   & 자동차[dzado\textipa{N}ts\super{h}a], 전화[dz{\textturnv}nhwa], 짹짹[ts'\textipa{E}k{\textcorner}ts'\textipa{E}k{\textcorner}], \\
                                                   & 참새[ts\super{h}ams\super{h}\textipa{E}], 책상[ts\super{h}\textipa{E}k{\textcorner}s'a\textipa{N}], 코끼리[k\super{h}ok{\textcorner}iri], \\
                                                   & 토끼[t\super{h}ok{\textcorner}i], 풍선[p\super{h}u\textipa{N}s\super{h}{\textturnv}n], 호랑이[hora\textipa{N}i] \\
\hline
Additional test words                              & 강아지[ka\textipa{N}adzi], 김밥[kimbap{\textcorner}], 나비[nabi], 뚜껑 
                                                     [t'uk{\textcorner}{\textturnv}\textipa{N}], \\
                                                   & 라면[ramy{\textturnv}n], 버섯[p{\textturnv}s\super{h}{\textturnv}t{\textcorner}], 쓰레기[s'{\textturnm}regi], 양파[ya\textipa{N}p\super{h}a], \\
                                                   & 종이[tso\textipa{N}i], 캥거루[k\super{h}\textipa{E}\textipa{N}g{\textturnv}ru], 토마토[t\super{h}omat\super{h}o], \\
                                                   & 헬리콥터[he{\textturny\textturny}ik\super{h}op{\textcorner}t\super{h}{\textturnv}] \\
\hline

\end{tabular}
\caption{Selected words list.}
\label{tab:selected_words}
\end{table}

As shown in Table \ref{tab:selected_words}, the analysed data included 73 Korean words, in two groups of 37 and 30 words with 6 overlapping words between each test tool. These word groups were selected from the Assessment of Phonology and Articulation for Children \cite{kim2007assessment} and the Urimal Test of Articulation and Phonology \cite{kim2004urimal}. A further set of 12 words (additional test words) was selected to enable to thorough assessment of the model’s performance for the evaluation of all Korean phonemes across all locations (word-initial, word-medial and word-final).

The participants were 137 children between 30 and 89 months old who visited the outpatient clinic of the Department of Paediatric Rehabilitation Medicine for suspected SSDs due to inaccurate pronunciation. Exclusion criteria were children with difficulty cooperating with the assessment, low speech rates that made transcribing their answers challenging, residence experience in another country and significant dialectal variation. Using a recorder (Tascam DR 44-WL), the participants were documented naming the object in a picture after looking at it and if they found it difficult to say the answer, the tester spoke first and encouraged the participant to imitate the utterance. 

\begin{table}[ht]
    \renewcommand{\arraystretch}{1.2} 
    \centering
    \begin{minipage}[b]{0.45\linewidth} 
        \centering
        \begin{tabular}{lcc}
        \hline
        \textbf{Data Type} & \textbf{Speakers} & \textbf{Duration (Hours)} \\
        \hline
        Train & 95 & 1.56 \\
        Dev & 12 & 0.19 \\
        Test & 30 & 0.54 \\
        \hline
        \end{tabular}
        \caption{Dataset information.}
        \label{tab:dataset_info}
    \end{minipage}
    \hfill
    \begin{minipage}[b]{0.5\linewidth} 
        \centering
        \begin{tabular}{ll}
        \hline
        \textbf{Stage}                 & \textbf{Transcript}           \\ \hline
        1. Target word in recording      & 싸움[s'aum]              \\
        2. Transcribed word              & 짜움[ts'aum]             \\
        3. Separated word for training   & ㅉㅏㅇㅜㅁ[ts'aum]             \\ \hline
        \end{tabular}
        \caption{Example in pre-processing.}
        \label{tab:example_pre}
    \end{minipage}
\end{table}

As shown in Table \ref{tab:dataset_info}, the final data consisted of the speech of the 137 participants who uttered all 73 words. The datasets were separated into training, validation and test sets to avoid the overlapping of speakers between sets. This study was conducted in accordance with the Declaration of Helsinki and approved by the Institutional Review Board of the Catholic University of Korea, Incheon St. Mary’s Hospital (protocol code OC22OISI0041; received approval on August 19, 2022).

We use target texts for speech made by speech-language pathologists who annotate the utterances as pronounced instead of using the original spelling of the words. Following this, for a detailed analysis of the results of the recognition, the consonants and vowels of the target text were separated for fine-tuning (see Table \ref{tab:example_pre}). The sample rate for each audio recording was converted to 16000 for training.

\subsection{Model fine-tuning}
We selected the model \href{https://huggingface.co/facebook/wav2vec2-xls-r-1b}{wav2vec2-xls-r-1b} as a pre-trained model due to its advantages in GPU memory size. In the fine-tuning stage, we used an NVIDIA Tesla V100 (32 GB) GPU. In batch size 16, we set the gradient accumulation step to 2 and the epoch to 30, using the Adam optimiser with an initial learning rate of 5 × 10-5. Evaluation was performed every 400 steps, with a warm-up phase of 500 steps. The vocabulary consisted of 40 Korean characters (consonants and vowels), with five tokens required for training and decoding in the wav2vec2.0 framework (blank space, unknown token, padding token, beginning-of-sentence and end-of-sentence).

\subsection{LM weight and decoder}
In this experiment, we explored the use of LM weighting as well as the proper timing for the training set. First, to determine how LM weighting affects decoding performance, we conducted fine-tuning with a 1-hour train dataset, estimating the recognition performance. After completing our analysis, we verified that the application of LM weights enhanced recognition performance, even when the model was trained with the entire dataset.

\begin{table}[h]
\renewcommand{\arraystretch}{1.2} 
\centering
\small 
\begin{tabular}{p{0.15\textwidth}p{0.15\textwidth}p{0.15\textwidth}p{0.15\textwidth}p{0.15\textwidth}}
\hline
\textbf{Origin Word} & \textbf{Main Category} & \textbf{Middle Category} & \textbf{Subcategory} & \textbf{Applied Word} \\
\hline
호랑이[hora\textipa{N}i]         & Word error patterns & Deletion     & Word-medial  & 호라이[horai] \\
                               &                     &              & coda deletion&             \\
\hline
단추[tants\super{h}u] & Segmental           & Manner of    & Plosive & 단뚜[tant'u],\\
                               & phoneme changes     & articulation &         & 단두[tantu], \\
                               &                     &              &         & 단투[tant\super{h}u] \\                              
\hline
바지[padzi]                     & Distortion errors   & Place of     & Lip rounding & 바쥐[padzwi],\\
                               &                     & articulation &              &  봐지[pwadzi], \\
                               &                     &              &              & 봐쥐[pwadzwi] \\
\hline
\end{tabular}
\caption{Example of generating words in pronunciation error dictionary.}
\label{tab:pronun}
\end{table}

To compare whether LM weighting improved performance, for use as an LM, we generated a pronunciation error dictionary composed of several cases of phoneme sequences that may occur when words are pronounced incorrectly (see Table \ref{tab:phone_seq}). Taking into account the phonological phenomena and characteristics for each word, the dictionary consists of phoneme sequences that are only applied to one error type, two consecutive error types or pronunciation errors occurring in the actual clinical field Table \ref{tab:pronun}. The LM was generated in a 5-gram ARPA format using \href{https://github.com/kpu/kenlm}{KenLM}.

A decoder is a layer in the ASR network joining the input speech into the appropriate chains of words or characters. Wav2vec2.0 employs the Connectionist Temporal Classification model \cite{graves2006connectionist} as a decoder to recognise words as they are pronounced, regardless of errors in spelling accuracy \cite{huggingface_asr_model}. We determined that this decoder was suitable for our task, which required the recognition of literal utterances and the identification of spelling errors.

\subsection{Performance evaluation}
Because we used the human-annotated text of utterances as the target text in model training, we considered it unnecessary to measure F1 scores as in other mispronunciation detection tasks that use words with original spellings as target words. Instead, we use an accuracy metric to measure how well the model-predicted text matches the target text annotated by pathologists. Because the model training was conducted through separating consonants and vowels, we measured accuracy with an error rate to compare differences in prediction texts from targeting one per consonant and vowel. This error is called the Phoneme Error Rate (PER) because each consonant and each vowel are considered a separate phoneme here. To calculate the error rate, we use the script available via \href{https://github.com/kaldi-asr/kaldi}{GitHub}. In evaluation, we removed the punctuation marks from the model output before calculating the error rate.

\section{Results}

\begin{table}[h]
\renewcommand{\arraystretch}{1.2}
\centering
\begin{tabular}{llcc}
\hline
\textbf{Pre-trained Model} & \textbf{Fine-tuning Time} & \textbf{LM Used} & \textbf{Phoneme Error Rate} \\ \hline
Whisper-large              & only pre-trained          & False            & 49.83                       \\
Whisper-large-v2           & only pre-trained          & False            & 46.80                       \\
wav2vec2-xls-r-1b          & 1 hour                    & False            & 11.25                       \\
wav2vec2-xls-r-1b          & 1 hour                    & True             & 11.25                       \\
wav2vec2-xls-r-1b          & 1.56 hours                & False            & \textbf{10.25}                       \\
wav2vec2-xls-r-1b          & 1.56 hours                & True             & 10.54                       \\ \hline
\end{tabular}
\caption{Results of the ASR models in terms of the PER.}
\label{tab:performance}
\end{table}

The test results for Whisper \cite{radford2022robust} showed that Whisper-large and Whisper-large-v2 models had word error rates of 15.2\% and 14.3\%. Previous studies noted that the model was sufficiently robust to achieve sufficient recognition performance for dataset-specific tasks without fine-tuning. However, as shown in Table \ref{tab:performance}, in this experiment, both models had an error rate of about 50\% for the recognition of the speech of children with SSDs. 
However, in the case of the model fine-tuned with the wav2vec 2.0 pre-trained model, XLS-R \cite{babu2022xls}, for labelled speech data, a PER of 10\% was achieved. Furthermore, as the time spent by the training data increased, recognition performance was greater when the LM was not used than when it was.

\begin{table}[h]
\renewcommand{\arraystretch}{1.2}
\centering
\begin{tabular}{llccc}
\hline
\textbf{Pre-trained Model} & \textbf{Fine-tuning Time} & \textbf{LM Used} & \textbf{Consonant F1 Score} & \textbf{C-PER} \\ \hline
wav2vec2-xls-r-1b          & 1 hour                    & False            & 0.812             & 14.471         \\
wav2vec2-xls-r-1b          & 1 hour                    & True             & 0.812             & 14.776         \\
wav2vec2-xls-r-1b          & 1.56 hours                & False            & 0.812             & \textbf{13.429}         \\
wav2vec2-xls-r-1b          & 1.56 hours                & True             & \textbf{0.813}             & 13.67          \\ \hline
\end{tabular}
\caption{Results of the ASR models in terms of f1 score and PER of consonants.}
\label{tab:f1}
\end{table}

Because SSD children’s pronunciation is diagnosed by analysing the percentage of correct consonants \cite{shriberg1997percentage}, we obtain an f1 score for each model including only Korean consonants from the decoding results, as shown in \ref{tab:f1}. Among the models trained on full-time data, the model with LM weights had a higher F1 score, but this was not significantly different from the model without an LM weight. The C-PER value, which shows the error rate for only the phonemes that correspond to consonants, was the lowest for the model without an LM weight. In other words, the performance evaluation, taking into account the actual diagnosis, was not significantly different from the result for comparing the PER score to evaluate only the recognition performance of the ASR model.

\begin{figure}
  \centering
  \includegraphics[width=1\textwidth]{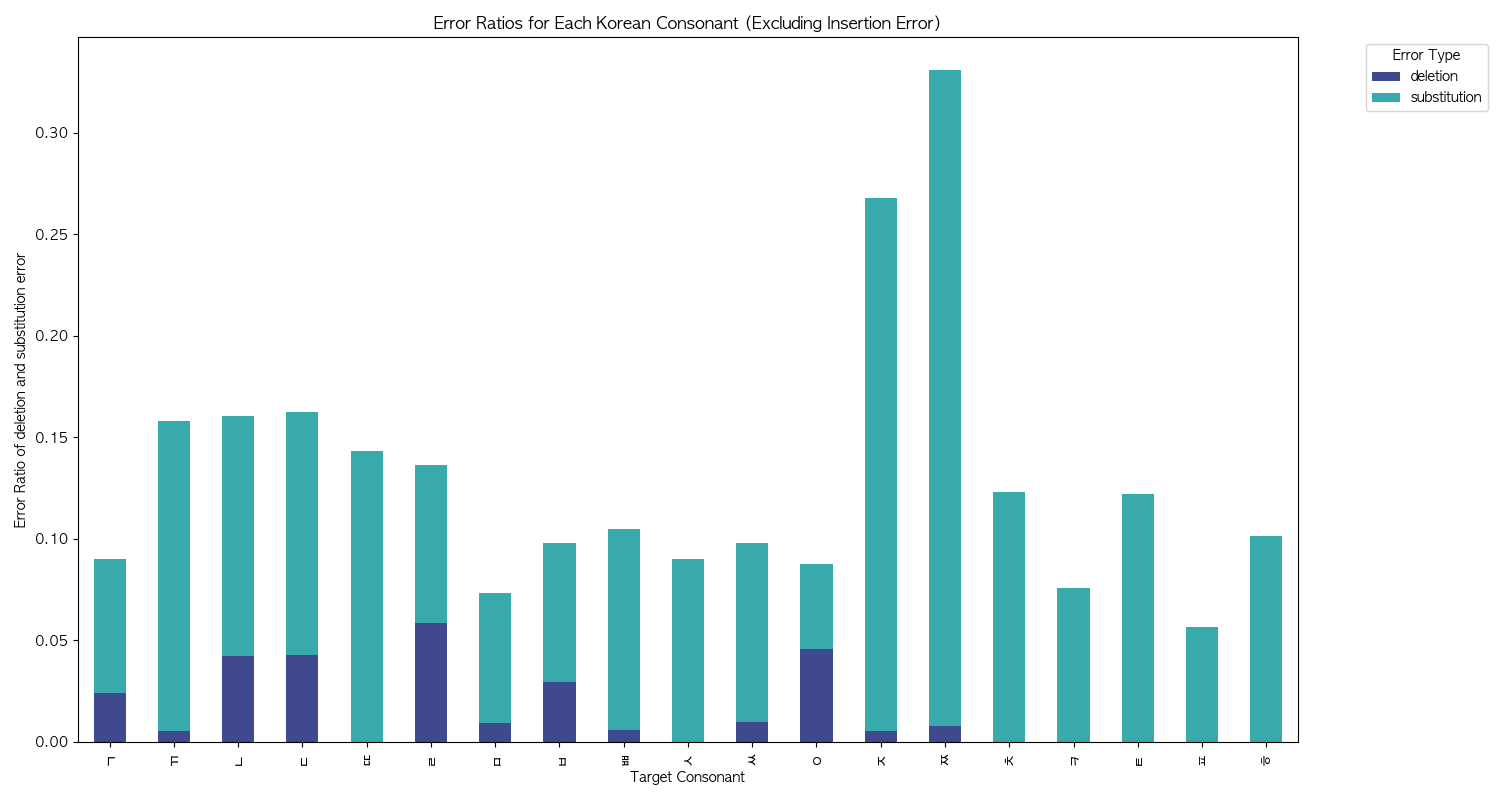}
  \caption{Error ratios for each Korean consonant (Excluding insertion error).}
  \label{fig:fig1}
\end{figure}

The bar chart presented in Figure \ref{fig:fig1} shows the percentage of errors in the recognition results for each target consonant. For speech recognition, each result can be decoded and assessed as correct (when the target and prediction are the same) or deletion, insertion, or substitution error. Because insertion errors are recognised when a new word or character is added in the absence of a target, those errors are not visible in the figure above, which calculates the error rate for the target words. The results show that the post-alveolar fricative consonants ‘ㅈ[dz]’ and ‘ㅉ[ts']’ have the highest error rate.

\begin{figure}
  \centering
  \includegraphics[width=1\textwidth]{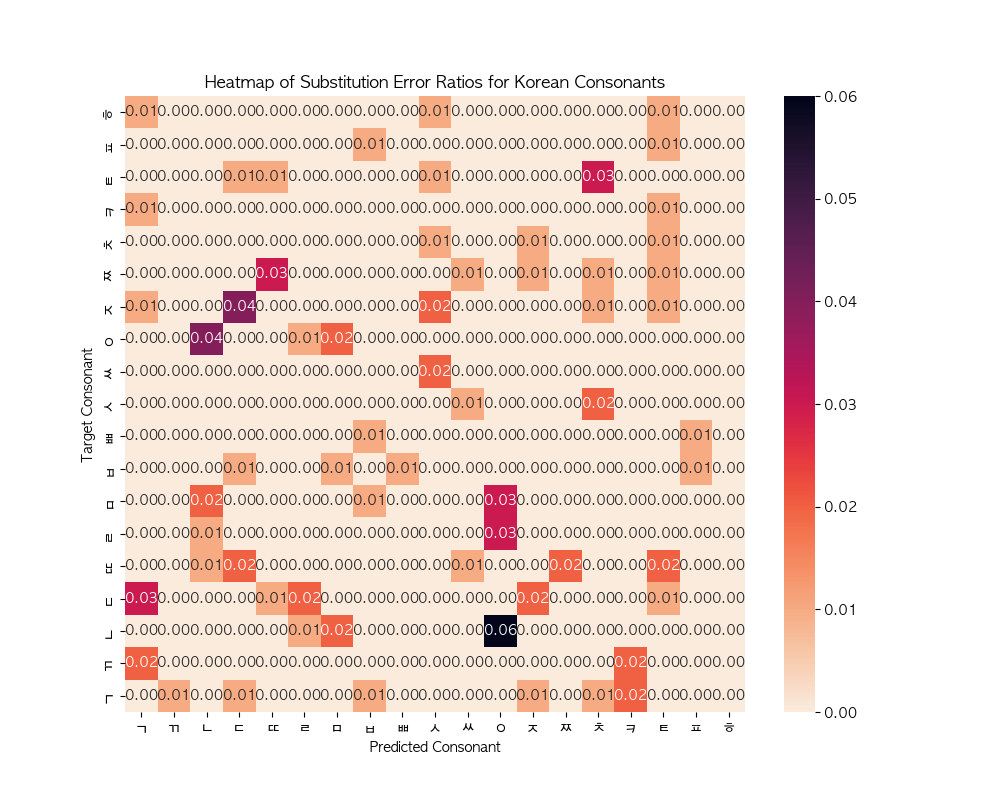}
  \caption{Heatmap of substitution error ratios for Korean consonants.}
  \label{fig:fig2}
\end{figure}

The heatmap chart in Figure \ref{fig:fig2} visualises the error percentage for each target-prediction consonant pair for all substitution errors. The main consonant pairs with higher error rates are mainly those where the target and predictor consonants have similar pronunciation, such as ‘ㄷ[d]’ and ‘ㅈ[dz]’ which are both articulated with the tongue near the alveolar ridge. We also found that ‘ㄴ’ was recognised as ‘ㅇ’ in the largest proportion of error cases, which may be due to the absence of a consonant in the initial sound of the word decoded as ‘ㅇ’ or because the two consonants are pronounced similarly in their coda sound ([n] and [\textipa{N}]), resulting in a substitution error.

\section{Discussion}
This study made several significant contributions to the development of an ASR model for the diagnosis of speech errors in children with SSDs. The main point obtained from the experiment is that only 1.5 hours of human-annotated speech was needed data to achieve the desired performance. We demonstrated that by fine-tuning the wav2vec2.0-based multilingual pre-trained model, we could build an ASR model that could recognise the speech of SSD children with a 90\% similarity to human-annotated text. This experiment demonstrated the feasibility with which an e2e-based ASR model can learn the acoustic features of speech of children with SSDs with a small amount of labelled audio data, even if the pronunciation is not perfect.

Most studies of mispronunciation have focused on identifying the pronunciation errors caused by second language learners when speaking a non-native language. This study, by contrast with those, has the great advantage of training the ASR model in the detection of pronunciation errors due to SSDs even in the speaker’s native language. In addition, this experiment features significant contributions that enable a high accuracy of recognition of non-English speech of SSDs. It was possible to make these contributions because speech recognition in non-English languages is less well-researched than in English. In particular, it should also be noted that speech recognition for children with SSDs has not been a research priority. These aspects allowed us to fill the technological gaps by building a model for diagnosing pronunciation errors in non-English speech using a state-of-the-art e2e-based ASR model.

We also find a necessity for LM weighting for downstream ASR tasks featuring low levels of resource data. In this experiment, models without LM weighting achieved greater recognition accuracy than the weighted model, although the LM was built by predicting the various pronunciation errors that SSDs could commonly make. LM weighting is often used when it is necessary for a model to decode speech to text with the intended grammatical or lexical order. Taking the usage of LM into account, this method may not be appropriate for this task, which requires the sounds of a single word to be learned as they are pronounced. As the task was focused on single word recognition, there was no necessity to learn grammatical rules, nor was there a requirement to decode specific words for projection. Additionally, while LMs are typically employed in scenarios that feature insufficient data for learning acoustic features, the self-supervised learning approach in wav2vec2.0 likely provided sufficient training data to adequately learn the acoustic features in this task.

Nevertheless, this study had certain limitations. Particularly notable were the decoding errors in the results, especially among consonants that have similar manners and places of articulation. As indicated in Figure \ref{fig:fig2}, the consonants ‘ㄷ[d]’ and ‘ㅈ[dz]’ were frequently confused in the utterance decoding. These errors stemmed not only from the model’s own decoding inaccuracies but also from challenges in correlating the acoustic features with the phonemes in the vocabulary items, which is referred to as ‘non-categorical phonemes’ \cite{mao2018unsupervised}. The primary goal of this study was streamlining the diagnostic process, so the training pipeline did not incorporate pronunciation symbols. However, for tasks that demanded precise recognition for these non-categorical phonemes, future research should consider developing a dataset including pronunciation symbols.

Additionally, while this study concentrated on the creation of an ASR model for diagnostic purposes, further investigation is needed to explore the correlation between consonant groups prone to errors in recognition results and the actual utterances of children with SSDs. While we were able to calculate the error rates for each consonant depicted in Figure \ref{fig:fig1}, we did not extend our examination to include a pathological analysis of these results, instead focusing on the primary goal of the study.

In addition, several factors must be taken into account before receiving clinical application. The factors of unexpected ambient noise or unclear pronunciation can impact diagnostic accuracy \cite{chun2023development}. Because this experiment did not account for such exceptional cases in the training of the model or in the measurement of recognition accuracy, future studies should focus on the development of a model that demonstrates robustness in real-world environments.

This research indicated the feasibility of constructing ASR models to achieve up to 90\% of level of performance level of human evaluators, without requiring extensive data or intricate procedures for alignment with the linguistic demands needed for diagnosing SSDs. The adoption of models developed through this method promises to alleviate the labour-intensive aspects of linguistic treatment and rehabilitation. Overall, this study made significant contributions to the field, paving the way for future research to develop optimal ASR models for identifying speech errors in children with SSDs.

\section{Conclusion}
In this study, we demonstrated the effectiveness of a fine-tuned, large-scale, multilingual pre-trained model for evaluating the pronunciation of native Korean-speaking children with SSDs. With an approximate 90\% accuracy, our model was able to recognise speech from children with SSDs in 73 Korean words, with only about 1.5 hours of training data and without the use of an LM. This indicates that adequate accuracy is attainable in training speech recognition models to recognise specified words, even with a low-resource dataset, to identify speech errors. While further research can be conducted to refine the model’s ability to reduce misrecognitions, our experimental results indicate the potential for simplifying the complex manual procedures currently used in diagnosing speech errors in Korean-speaking children in clinical settings through the application of ASR technology.

\section*{Funding}
This work was supported by the National Research Foundation of Korea (NRF) grant funded by the Korea government (MSIT) (No. NRF-2022R1C1C1008337)

\section*{Disclosure of interest}
The authors report no conflict of interest.

\section*{Data availability statement}
The speech datasets generated and/or analysed during the current study are not publicly available because no prior consent was obtained from the participants to share their data. However, interested researchers may request access to the other data from the corresponding author\footnote{dhjangmd@naver.com}, subject to appropriate ethical considerations and approval.

\bibliographystyle{unsrtnat}  
\bibliography{references}

\begin{thebibliography}{30}
\providecommand{\natexlab}[1]{#1}
\providecommand{\url}[1]{\texttt{#1}}
\expandafter\ifx\csname urlstyle\endcsname\relax
  \providecommand{\doi}[1]{doi: #1}\else
  \providecommand{\doi}{doi: \begingroup \urlstyle{rm}\Url}\fi

\bibitem[Dudy et~al.(2015)Dudy, Asgari, and Kain]{dudy2015pronunciation}
Shimon Dudy, Meysam Asgari, and Alexander Kain.
\newblock Pronunciation analysis for children with speech sound disorders.
\newblock In \emph{Annual International Conference of the IEEE Engineering in Medicine and Biology Society}, pages 5573--5576. IEEE, 2015.
\newblock \doi{10.1109/EMBC.2015.7319655}.
\newblock URL \url{https://www.ncbi.nlm.nih.gov/pmc/articles/PMC4710861/}.

\bibitem[Dudy et~al.(2018)Dudy, Bedrick, Asgari, and Kain]{dudy2018automatic}
Shimon Dudy, Steven Bedrick, Meysam Asgari, and Alexander Kain.
\newblock Automatic analysis of pronunciations for children with speech sound disorders.
\newblock \emph{Computer Speech \& Language}, 50:\penalty0 62--84, 2018.
\newblock \doi{10.1016/j.csl.2017.12.006}.
\newblock URL \url{https://doi.org/10.1016/j.csl.2017.12.006}.

\bibitem[Povey et~al.(2011)Povey, Ghoshal, Boulianne, Burget, Glembek, Goel, Hannemann, Motlicek, Qian, Schwarz, et~al.]{povey2011kaldi}
Daniel Povey, Arnab Ghoshal, Gilles Boulianne, Lukas Burget, Ondrej Glembek, Nagendra Goel, Mirko Hannemann, Petr Motlicek, Yanmin Qian, Petr Schwarz, et~al.
\newblock The kaldi speech recognition toolkit.
\newblock In \emph{IEEE 2011 workshop on automatic speech recognition and understanding}. IEEE Signal Processing Society, 2011.

\bibitem[Chorowski et~al.(2015)Chorowski, Bahdanau, Serdyuk, Cho, and Bengio]{chorowski2015attention}
Jan~K Chorowski, Dzmitry Bahdanau, Dmitriy Serdyuk, Kyunghyun Cho, and Yoshua Bengio.
\newblock Attention-based models for speech recognition.
\newblock In C.~Cortes, N.~Lawrence, D.~Lee, M.~Sugiyama, and R.~Garnett, editors, \emph{Advances in Neural Information Processing Systems}, volume~28. Curran Associates, Inc., 2015.
\newblock URL \url{https://proceedings.neurips.cc/paper_files/paper/2015/file/1068c6e4c8051cfd4e9ea8072e3189e2-Paper.pdf}.

\bibitem[Bahdanau et~al.(2016)Bahdanau, Chorowski, Serdyuk, Brakel, and Bengio]{bahdanau2016end}
Dzmitry Bahdanau, Jan Chorowski, Dmitriy Serdyuk, Philémon Brakel, and Yoshua Bengio.
\newblock End-to-end attention-based large vocabulary speech recognition.
\newblock In \emph{2016 IEEE International Conference on Acoustics, Speech and Signal Processing (ICASSP)}, pages 4945--4949, 2016.
\newblock \doi{10.1109/ICASSP.2016.7472618}.

\bibitem[Chiu et~al.(2018)Chiu, Sainath, Wu, Prabhavalkar, Nguyen, Chen, Kannan, Weiss, Rao, Gonina, Jaitly, Li, Chorowski, and Bacchiani]{chiu2018state}
Chung-Cheng Chiu, Tara~N. Sainath, Yonghui Wu, Rohit Prabhavalkar, Patrick Nguyen, Zhifeng Chen, Anjuli Kannan, Ron~J. Weiss, Kanishka Rao, Ekaterina Gonina, Navdeep Jaitly, Bo~Li, Jan Chorowski, and Michiel Bacchiani.
\newblock State-of-the-art speech recognition with sequence-to-sequence models.
\newblock In \emph{2018 IEEE International Conference on Acoustics, Speech and Signal Processing (ICASSP)}, pages 4774--4778, 2018.
\newblock \doi{10.1109/ICASSP.2018.8462105}.

\bibitem[Abate et~al.(2021)Abate, Tachbelie, and Schultz]{abate2021end}
Solomon~Teferra Abate, Martha~Yifiru Tachbelie, and Tanja Schultz.
\newblock End-to-end multilingual automatic speech recognition for less-resourced languages: The case of four ethiopian languages.
\newblock In \emph{ICASSP 2021 - 2021 IEEE International Conference on Acoustics, Speech and Signal Processing (ICASSP)}, pages 7013--7017, 2021.
\newblock \doi{10.1109/ICASSP39728.2021.9415020}.

\bibitem[Zuluaga-Gomez et~al.(2023)Zuluaga-Gomez, Prasad, Nigmatulina, Sarfjoo, Motlicek, Kleinert, Helmke, Ohneiser, and Zhan]{zuluaga2023pre}
Juan Zuluaga-Gomez, Amrutha Prasad, Iuliia Nigmatulina, Seyyed~Saeed Sarfjoo, Petr Motlicek, Matthias Kleinert, Hartmut Helmke, Oliver Ohneiser, and Qingran Zhan.
\newblock How does pre-trained wav2vec 2.0 perform on domain-shifted asr? an extensive benchmark on air traffic control communications.
\newblock In \emph{2022 IEEE Spoken Language Technology Workshop (SLT)}, pages 205--212. IEEE, 2023.

\bibitem[Xu et~al.(2022)Xu, Baevski, and Auli]{xu2022simple}
Qiantong Xu, Alexei Baevski, and Michael Auli.
\newblock {Simple and Effective Zero-shot Cross-lingual Phoneme Recognition}.
\newblock In \emph{Proc. Interspeech 2022}, pages 2113--2117, 2022.
\newblock \doi{10.21437/Interspeech.2022-60}.

\bibitem[Yi et~al.(2020)Yi, Wang, Cheng, Zhou, and Xu]{yi2021applying}
Cheng Yi, Jianzhong Wang, Ning Cheng, Shiyu Zhou, and Bo~Xu.
\newblock Applying wav2vec2. 0 to speech recognition in various low-resource languages.
\newblock \emph{arXiv preprint arXiv:2012.12121}, 2020.

\bibitem[Feng et~al.(2020)Feng, Fu, Chen, and Chen]{feng2020sed}
Yiqing Feng, Guanyu Fu, Qingcai Chen, and Kai Chen.
\newblock Sed-mdd: Towards sentence dependent end-to-end mispronunciation detection and diagnosis.
\newblock In \emph{ICASSP 2020 - 2020 IEEE International Conference on Acoustics, Speech and Signal Processing (ICASSP)}, pages 3492--3496, 2020.
\newblock \doi{10.1109/ICASSP40776.2020.9052975}.

\bibitem[Liu et~al.(2023)Liu, Wumaier, Wei, and Guo]{liu2023automatic}
Jiajun Liu, Aishan Wumaier, Dongping Wei, and Shen Guo.
\newblock Automatic speech disfluency detection using wav2vec2.0 for different languages with variable lengths.
\newblock \emph{Applied Sciences}, 13\penalty0 (13):\penalty0 7579, 2023.
\newblock \doi{10.3390/app13137579}.
\newblock URL \url{https://doi.org/10.3390/app13137579}.

\bibitem[Peng et~al.(2021)Peng, Fu, Lin, Ke, and Zhan]{peng2021study}
Linkai Peng, Kaiqi Fu, Binghuai Lin, Dengfeng Ke, and Jinsong Zhan.
\newblock {A Study on Fine-Tuning wav2vec2.0 Model for the Task of Mispronunciation Detection and Diagnosis}.
\newblock In \emph{Proc. Interspeech 2021}, pages 4448--4452, 2021.
\newblock \doi{10.21437/Interspeech.2021-1344}.

\bibitem[Peng et~al.(2023)Peng, Gao, Bao, Li, and Zhang]{peng2023end}
Linkai Peng, Yingming Gao, Rian Bao, Ya~Li, and Jinsong Zhang.
\newblock End-to-end mispronunciation detection and diagnosis using transfer learning.
\newblock \emph{Applied Sciences}, 13\penalty0 (11):\penalty0 6793, 2023.
\newblock \doi{10.3390/app13116793}.
\newblock URL \url{https://doi.org/10.3390/app13116793}.

\bibitem[Yang et~al.(2022)Yang, Hirschi, Looney, Kang, and Hansen]{yang2022improving}
Mu~Yang, Kevin Hirschi, Stephen~Daniel Looney, Okim Kang, and John~H.L. Hansen.
\newblock {Improving Mispronunciation Detection with Wav2vec2-based Momentum Pseudo-Labeling for Accentedness and Intelligibility Assessment}.
\newblock In \emph{Proc. Interspeech 2022}, pages 4481--4485, 2022.
\newblock \doi{10.21437/Interspeech.2022-11039}.

\bibitem[Lee et~al.(2023)Lee, Lee, Bong, Yoo, and Kim]{lee2022end}
Jung~Hyuk Lee, Geon~Woo Lee, Guiyoung Bong, Hee~Jeong Yoo, and Hong~Kook Kim.
\newblock End-to-end model-based detection of infants with autism spectrum disorder using a pretrained model.
\newblock \emph{Sensors}, 23\penalty0 (1):\penalty0 202, 2023.
\newblock \doi{10.3390/s23010202}.
\newblock URL \url{https://doi.org/10.3390/s23010202}.

\bibitem[Escobar-Grisales et~al.(2023)Escobar-Grisales, Ríos-Urrego, and Orozco-Arroyave]{escobargrisales2023deep}
Daniel Escobar-Grisales, Cristian~David Ríos-Urrego, and Juan~Rafael Orozco-Arroyave.
\newblock Deep learning and artificial intelligence applied to model speech and language in parkinson’s disease.
\newblock \emph{Diagnostics}, 13\penalty0 (13):\penalty0 2163, 2023.
\newblock \doi{10.3390/diagnostics13132163}.
\newblock URL \url{https://doi.org/10.3390/diagnostics13132163}.

\bibitem[Baevski et~al.(2020)Baevski, Zhou, Mohamed, and Auli]{baevski2020wav2vec}
Alexei Baevski, Yuhao Zhou, Abdelrahman Mohamed, and Michael Auli.
\newblock wav2vec 2.0: A framework for self-supervised learning of speech representations.
\newblock In H.~Larochelle, M.~Ranzato, R.~Hadsell, M.F. Balcan, and H.~Lin, editors, \emph{Advances in Neural Information Processing Systems}, volume~33, pages 12449--12460. Curran Associates, Inc., 2020.
\newblock URL \url{https://proceedings.neurips.cc/paper_files/paper/2020/file/92d1e1eb1cd6f9fba3227870bb6d7f07-Paper.pdf}.

\bibitem[Conneau et~al.(2021)Conneau, Baevski, Collobert, Mohamed, and Auli]{conneau2021unsupervised}
Alexis Conneau, Alexei Baevski, Ronan Collobert, Abdelrahman Mohamed, and Michael Auli.
\newblock {Unsupervised Cross-Lingual Representation Learning for Speech Recognition}.
\newblock In \emph{Proc. Interspeech 2021}, pages 2426--2430, 2021.
\newblock \doi{10.21437/Interspeech.2021-329}.

\bibitem[Babu et~al.(2022)Babu, Wang, Tjandra, Lakhotia, Xu, Goyal, Singh, {von Platen}, Saraf, Pino, Baevski, Conneau, and Auli]{babu2022xls}
Arun Babu, Changhan Wang, Andros Tjandra, Kushal Lakhotia, Qiantong Xu, Naman Goyal, Kritika Singh, Patrick {von Platen}, Yatharth Saraf, Juan Pino, Alexei Baevski, Alexis Conneau, and Michael Auli.
\newblock {XLS-R: Self-supervised Cross-lingual Speech Representation Learning at Scale}.
\newblock In \emph{Proc. Interspeech 2022}, pages 2278--2282, 2022.
\newblock \doi{10.21437/Interspeech.2022-143}.

\bibitem[Getman et~al.(2022)Getman, Al-Ghezi, Voskoboinik, Grósz, Kurimo, Salvi, Svendsen, and Strömbergsson]{getman2022wav2vec2}
Yaroslav Getman, Ragheb Al-Ghezi, Katja Voskoboinik, Tamás Grósz, Mikko Kurimo, Giampiero Salvi, Torbjørn Svendsen, and Sofia Strömbergsson.
\newblock {wav2vec2-based Speech Rating System for Children with Speech Sound Disorder}.
\newblock In \emph{Proc. Interspeech 2022}, pages 3618--3622, 2022.
\newblock \doi{10.21437/Interspeech.2022-10103}.

\bibitem[Radford et~al.(2023)Radford, Kim, Xu, Brockman, McLeavey, and Sutskever]{radford2022robust}
Alec Radford, Jong~Wook Kim, Tao Xu, Greg Brockman, Christine McLeavey, and Ilya Sutskever.
\newblock Robust speech recognition via large-scale weak supervision.
\newblock In \emph{International Conference on Machine Learning}, pages 28492--28518. PMLR, 2023.

\bibitem[Jain et~al.(2023)Jain, Barcovschi, Yiwere, Corcoran, and Cucu]{jain2023adaptation}
Rishabh Jain, Andrei Barcovschi, Mariam Yiwere, Peter Corcoran, and Horia Cucu.
\newblock Adaptation of whisper models to child speech recognition, 2023.

\bibitem[Kim et~al.(2007)Kim, Pae, and Park]{kim2007assessment}
Min~Jung Kim, Soyeong Pae, and Chang~Il Park.
\newblock Assessment of phonology and articulation for children (apac), 2007.

\bibitem[Kim et~al.(2004)Kim, Shin, Kim, and Ha]{kim2004urimal}
YT~Kim, MJ~Shin, SJ~Kim, and JW~Ha.
\newblock Urimal test of articulation and phonology (u-tap), 2004.

\bibitem[Graves et~al.(2006)Graves, Fernández, Gomez, and Schmidhuber]{graves2006connectionist}
Alex Graves, Santiago Fernández, Faustino Gomez, and Jürgen Schmidhuber.
\newblock Connectionist temporal classification: labelling unsegmented sequence data with recurrent neural networks.
\newblock In \emph{Proceedings of the 23rd international conference on Machine learning (ICML '06)}, pages 369--376, New York, NY, USA, 2006. Association for Computing Machinery.
\newblock \doi{10.1145/1143844.1143891}.
\newblock URL \url{https://doi.org/10.1145/1143844.1143891}.

\bibitem[Face(2023)]{huggingface_asr_model}
Hugging Face.
\newblock Chapter 5: Asr model - hugging face audio course.
\newblock \url{https://huggingface.co/learn/audio-course/chapter5/asr_model}, 2023.
\newblock Accessed: 2024-03-10.

\bibitem[Shriberg et~al.(1997)Shriberg, Austin, Lewis, McSweeny, and Wilson]{shriberg1997percentage}
Lawrence~D Shriberg, Diane Austin, Barbara~A Lewis, Jane~L McSweeny, and David~L Wilson.
\newblock The percentage of consonants correct (pcc) metric: Extensions and reliability data.
\newblock \emph{Journal of Speech, Language, and Hearing Research}, 40\penalty0 (4):\penalty0 708--722, 1997.

\bibitem[Mao et~al.(2018)Mao, Li, Li, Wu, Liu, and Meng]{mao2018unsupervised}
Shaoguang Mao, Xu~Li, Kun Li, Zhiyong Wu, Xunying Liu, and Helen Meng.
\newblock Unsupervised discovery of an extended phoneme set in l2 english speech for mispronunciation detection and diagnosis.
\newblock In \emph{2018 IEEE International Conference on Acoustics, Speech and Signal Processing (ICASSP)}, pages 6244--6248, 2018.
\newblock \doi{10.1109/ICASSP.2018.8462635}.

\bibitem[Chun et~al.(2023)Chun, Park, Ryu, and Jang]{chun2023development}
Seok-Joo Chun, Jung~Bin Park, Hyejo Ryu, and Bum-Sup Jang.
\newblock Development and benchmarking of a korean audio speech recognition model for clinician-patient conversations in radiation oncology clinics.
\newblock \emph{International Journal of Medical Informatics}, 176:\penalty0 105112, 2023.
\newblock \doi{10.1016/j.ijmedinf.2023.105112}.
\newblock URL \url{https://pubmed.ncbi.nlm.nih.gov/37276615/}.

\end{thebibliography}

\clearpage
\begin{table}[h]
\renewcommand{\arraystretch}{1.2}
\centering
\begin{tabular}{lll}
\hline
\textbf{Main category}    & \textbf{Middle category}          & \textbf{Subcategory} \\
\hline
Word error patterns       & Deletion                          & Syllable deletion    \\
                          &                                   & Word-initial consonant deletion \\
                          &                                   & Word-medial consonant deletion \\
                          &                                   & Liquid deletion \\
                          &                                   & Word-medial coda deletion \\
                          &                                   & Word-final coda deletion \\
                          & Insertion                         & Syllable insertion   \\
                          &                                   & Vowel insertion      \\
                          &                                   & Consonant insertion  \\
                          & Reduplication                     & Syllable reduplication \\ 
                          &                                   & Vowel harmony        \\
                          &                                   & Consonant harmony    \\
                          & Transposition and Migration       & Syllable transposition \\
                          &                                   & Phoneme transposition \\
                          & Consonant cluster simplification  & Typical consonant cluster simplification \\
                          &                                   & Atypical consonant cluster simplification \\
\hline
Segmental phoneme changes & Place of articulation             & Fronting \\
                          &                                   & Bilabial \\
                          &                                   & Alveolo-Palatal \\
                          &                                   & Velar \\
                          &                                   & Glottal \\
                          & Manner of articulation            & Gliding \\   
                          &                                   & Denasalisation \\
                          &                                   & Nasal \\
                          &                                   & Plosive \\
                          &                                   & Affricative \\
                          &                                   & Fricative \\
                          &                                   & Stopping of liquid \\
                          &                                   & Liquid simplification \\
                          & Phonation Types                   & Tense \\   
                          &                                   & Lax \\
                          &                                   & Aspiration \\
                          &                                   & Unaspiration \\
\hline
Distortion errors         & Place of articulation             & Lip rounding \\
                          &                                   & Labiodental \\
                          &                                   & Interdental \\
                          &                                   & Tongue tip \\
                          &                                   & Palatal \\
\hline
\end{tabular}
\caption{Case of phoneme sequences for pronunciation error dictionary.}
\label{tab:phone_seq}
\end{table}

\clearpage
\vspace*{-\topskip}
\vspace*{-1in}
\vspace*{-\headsep}
\vspace*{-\headheight}
\vspace*{-\topmargin}

\clearpage
\vspace*{-\topskip}
\vspace*{-1in}
\vspace*{-\headsep}
\vspace*{-\headheight}
\vspace*{-\topmargin}

\end{document}